\documentclass[11pt,a4paper]{article}
\usepackage{amsmath,amssymb}
\usepackage[hyperref]{emnlp2020}
\usepackage{cleveref}
\usepackage{times}
\usepackage{latexsym}

\usepackage{microtype}

\usepackage{url}
\usepackage{graphicx}
\usepackage{subcaption}
\usepackage{amsfonts}
\usepackage{multirow}
\usepackage{makecell}
\usepackage{mathrsfs}

\aclfinalcopy 

\usepackage{adjustbox}
\usepackage{booktabs}
\usepackage{footnote}
\makesavenoteenv{tabular}
\makesavenoteenv{table}

\title{The Ubiqus English-Inuktitut System for WMT20}

\author{François Hernandez \\
  Ubiqus, Paris, France \\
  \texttt{fhernandez@ubiqus.com} \\\And
  Vincent Nguyen \\
  Ubiqus, Paris, France \\
  \texttt{vnguyen@ubiqus.com} \\}

\date{}

\begin{document}
\maketitle
\begin{abstract}
This paper describes Ubiqus' submission to the WMT20 English-Inuktitut shared news translation task. Our main system, and only submission, is based on a multilingual approach, jointly training a Transformer model on several agglutinative languages. The English-Inuktitut translation task is challenging at every step, from data selection, preparation and tokenization to quality evaluation down the line. Difficulties emerge both because of the peculiarities of the Inuktitut language as well as the low-resource context.

\end{abstract}

\section{Introduction}

Ubiqus participated in the English to Inuktitut news translation task of WMT20. We performed a single submission, based on an unconstrained multilingual setup. The approach consists of jointly training a traditional Transformer \cite{NIPS2017_7181} model on several agglutinative languages in order to benefit from them for the low-resource English-Inuktitut task \cite{aharoni-etal-2019-massively}.

Though the dataset provided for the task is sizable, with more than a million segments, it's quite narrow domain-wise, as it comes from proceedings of the Nunavut Hansard. The task being translation of news, it's expected to be a much wider domain. For that purpose, we extended the task with datasets of other - linguistically near - languages, as well as in-house datasets introducing more diversity to the domain.

All experiments were performed with the OpenNMT \cite{klein-etal-2017-opennmt} toolkit, with \emph{Tokenizer} \footnote{https://github.com/OpenNMT/Tokenizer} for data preprocessing and \emph{OpenNMT-py} \footnote{https://github.com/OpenNMT/OpenNMT-py} for model training and inference.

\section{Data}

\subsection{Training corpora}

Based on prior internal work on English-Inuktitut translation tasks as well as other low-resource tasks, we focused our experiments on multilingual setups. Inuktitut is an agglutinative language, with a lot of particularities. Some \emph{Uralic languages} like Finnish and Estonian can be considered close to Inuktitut in some linguistic aspects.

Most of our experiments are \emph{unconstrained} with regards to the original WMT task in three ways:
\begin{itemize}
    \setlength\itemsep{.1em}
    \item some datasets are taken from previous WMT tasks (English-Finnish, English-Estonian);
    \item some datasets are not in the WMT scope (more recent ParaCrawl\footnote{https://paracrawl.eu} versions);
    \item some datasets were built in-house at Ubiqus Labs.
\end{itemize}

Some Inuktitut resources can easily be found on the internet, mostly from official government of Nunavut websites and initiatives. We performed two sets of data retrieval: a first one based on parallel crawling of multilingual websites, and a second one based on manual retrieval of parallel documents (mostly in PDF format) which then were automatically aligned with a commercial tool. In prior experiments, we also built a set of parallel news articles. Articles were manually retrieved and aligned from both the \emph{Inuktitut} \footnote{https://www.itk.ca/category/inuktitut-magazine/} magazine, which provides parallel versions of all its content in English, French, Inuktitut and Inuinnaqtun, and the Nunatsiaq News\footnote{https://nunatsiaq.com} website, which provides part of its content in both Inuktitut and English. We decided not to include this last dataset because of its high proximity with the \emph{newsdev2020} and \emph{newstest2020} of the task.

A summary of all the datasets used in the experiments is available in Table \ref{datasets}.

\begin{table*}
\centering
\begin{tabular}{llll}
\toprule 
\textbf{Dataset} & \textbf{Origin} & \textbf{Raw} & \textbf{Selected}\\
\midrule 
Nunavut Hansard v3.0 \cite{joanis-etal-2020-nh30} & WMT EN-IU 2020 Task & 2,550,682 & 737,375 \\
\midrule 
$\star$Europarl English-Finnish & WMT EN-FI 2019 Task & 1,969,624 & 1,564,994  \\
$\star$Europarl English-Estonian & WMT EN-ET 2018 Task & 651,236 & 566,815  \\
\midrule 
$\star$ParaCrawlv6 English-Finnish & ParaCrawl Project & 4,286,642 & 4,207,262 \\
$\star$ParaCrawlv6 English-Estonian & ParaCrawl Project & 1,785,161 & 1,755,013 \\
\midrule 
$\star$Public Documents & Ubiqus & 102,567 & 66,159 \\
$\star$Public Websites & Ubiqus & 2,035,594 & 31,025 \\
\bottomrule 
\end{tabular}
\caption{\label{datasets}
Characteristics of the datasets used in the experiments. Datasets marked with $\star$ are considered out of the constraints of the WMT English-Inuktitut task.
}
\end{table*}

\subsection{Evaluation sets}

During our experiments, we conducted evaluation of the trained models with the provided \emph{newsdev2020-eniu} as well as the \emph{dev}, \emph{devtest} and \emph{test} parts of the Hansard dataset split. The latter were deduplicated prior to evaluation.

As a big part of our experiments revolve around multilingual aspects, we also used \emph{newstest2018-enfi} and \emph{newstest2019-enfi} for English-Finnish, as well as \emph{newstest2018-enet} for English-Estonian.

Finally, we also conducted some evaluation over the test part of our in-house dataset built from \emph{Inuktitut} magazine.

\subsection{Data selection and cleaning}

Deduplication as well as a few steps of cleaning were applied to every dataset. This consists of removing segments where:
\begin{itemize}
    \setlength\itemsep{.2em}
    \item average token length too short or too long;
    \item source is strictly equal to target;
    \item numbers do not match between source and target side;
    \item source to target character ratio is too extreme.
\end{itemize}

The difference between raw and selected dataset size is shown in Table \ref{datasets}. It is noticeable that this step is especially important for our in-house datasets, where automatically crawled and aligned data is particularly messy. Also, it seems the Nunavut Hansard dataset is quite clean but contains a lot of duplicates.

\subsection{Preprocessing and Tokenization}

We decided to work on romanized Inuktitut. This allows straightforward parameter and vocabulary sharing in a basic bilingual English-Inuktitut setup, as well as maximizing the potential benefits of parameter sharing in a multilingual setup. Hence, all the Inuktitut data was romanized prior to any other processing, and we only converted back our \emph{newstest2020-eniu} inferred hypothesis for submission.

All experiments were conducted on data tokenized with a BPE \cite{sennrich-etal-2016-neural} model with 12,000 merge operations, learned on the concatenation of all datasets -- both source and target -- presented in Table \ref{datasets} (without any particular sampling strategy). This leads to a final vocabulary size of approximately 14k tokens. The choice of a smaller number of BPE merge operations stems from the agglutinative aspect of the language, leading us to think that dividing long tokens into more subwords might be beneficial to learn and share more useful representations. This seems to be also the approach in the baseline system proposed in \cite{joanis-etal-2020-nh30}.

\section{Experiments}

\subsection{Mixing languages}

The method used to train models on multiple languages relies on the dataset weighting mechanism which is implemented within OpenNMT-py \cite{klein-etal-2020-opennmt}. When building batches, $weight_A$ examples are sampled from dataset $A$, then $weight_B$ from dataset $B$, and so on. This allows to dynamically subsample or oversample any specific dataset or language pair when training.

In order to allow Many-to-Many translation in a single shared model, we need to prepend each source with a tag indicating the target language \cite{johnson-etal-2017-googles}.

\subsection{Bilingual only}

Since we do not have any internal resource to assess the Inuktitut output, we started some bilingual experiments into English. With the English-Inuktitut datasets only, we realized that even with a base Transformer, the model converged very quickly and gave similar results with several varying hyper parameters. Also, changing the sampling weight of each sub-dataset did not have much impact to the final results. Moreover, English to Inuktitut bilingual experiments gave very poor results on our internal test set based on the Inuktitut Magazine. We hypothesize that there was some kind of overfitting to the Hansard domain. This is why we decided to extend a multilingual set up with more ``news'' based data.

\subsection{Multilingual}

We trained a few systems in the following order:
\begin{itemize}
    \item first, a bilingual (and bidirectional) English-Inuktitut system (base configuration Transformer) using the Nunavut dataset as well as our in-house Web and Documents datasets;
    \item next, we added the English-Finnish data;
    \item then, we added the English-Estonian data;
    \item finally, we increased the model size.
\end{itemize}

Results for these systems are summed up in Table \ref{scores}. We notice that multilingual setups are truly multilingual, in the sense that they provide output in the correct language, even though the scores are not very competitive (approx. 30\% below the best scores at the time of the corresponding WMT tasks).

We decided to retain the bigger model (medium Transformer) for the submission. Bigger multilingual models tend to be better with regards to human evaluation, probably because the tasks are better spread across the parameters. This can be a problem in case of overfitting, which does not seem to be the case here as the scores remain in the same range. Also, the bigger model seems to give marginally better results in the additional tasks (Finnish and Estonian), which leads us to think it will be more robust to new test sets. 

The configuration used for the final submission is the following:
\begin{itemize}
    \setlength\itemsep{.2em}
    \item \textbf{Corpora and weights}: shown in Table \ref{weights}.
    \item \textbf{Tokenization}: 12,000 BPE merge operations, learned on the concatenation of all datasets.
    \item \textbf{Model}: Transformer Medium (12 $heads$, $d_{model}=768$, $d_{ff}=3072$), with Relative Position Representations \cite{shaw-etal-2018-self}.
    \item \textbf{Training}: Trained with OpenNMT-py on 6 RTX 2080 Ti, using mixed precision. Initial batch size is around 50,000 tokens, final batch size around 200,000 tokens. Training was stopped at 100k steps. Averaging was done continuously through exponential moving average.
    \item \textbf{Inference}: Shown scores are obtained with beam search of size 5 and average length penalty. 
\end{itemize}

\begin{table}
\centering
\begin{tabular}{lc}
\toprule
Hansard & 15 \\
\midrule
$\star$Europarl en-et & 2  \\
$\star$Europarl en-fi & 2  \\
\midrule
$\star$ParaCrawlv6 en-et & 10 \\
$\star$ParaCrawlv6 en-fi & 10 \\
\midrule
$\star$Public Documents (Ubiqus) & 5 \\
$\star$Public Websites (Ubiqus) & 1 \\
\bottomrule
\end{tabular}
\caption{\label{weights}
Dataset weighting used for the submitted system.
}
\end{table}

\begin{table*}
\centering
\begin{adjustbox}{center}
\begin{tabular}{l|r|rrr|r|rr|r}
\toprule 
\textbf{System} & \textbf{nd20-eniu} & \textbf{dev} & \textbf{dev-test} & \textbf{test} & \textbf{IM} & \textbf{nt18-enfi} & \textbf{nt19-enfi} & \textbf{nt18-enet}\\
\midrule 
\cite{joanis-etal-2020-nh30} & - & 24.2 & 17.9 & 19.3 & - & - & - & - \\
\midrule 
en$\leftrightarrow$iu (base) & 15.6 & 23.9 & 17.7 & 19.4 & 4.7 & - & - & - \\
\midrule 
en$\leftrightarrow$iu/fi (base) & 15.6 & 23.6 & 17.5 & 19.2 & 7.6 & 11.8 & 16.3 & 2.4  \\
en$\leftrightarrow$iu/fi/et (base) & 15.5 & 23.3 & 17.4 & 18.9 & 7.6 & 11.9 & 16.6 & 16.9 \\
$\triangleright$ en$\leftrightarrow$iu/fi/et (medium) & 15.6 & 23.6 & 17.3 & 19.1 & 7.4 & 12.1 & 17.0 & 17.1 \\
\bottomrule 
\end{tabular}
\end{adjustbox}
\caption{\label{scores}
BLEU \cite{papineni-etal-2002-bleu} scores for our various experiments, obtained with SacreBLEU \cite{post-2018-call} v1.3.7. The submitted system is marked with $\triangleright$. \emph{dev}, \emph{dev-test} and \emph{test} refer to the Hansard dataset evaluation sets. IM stands for \emph{Inuktitut Magazine}.)
}

\end{table*}

\section{Future work}

Our experiments remain in a rather traditional Neural Machine Translation scope, with the only addition of multiple languages and dataset weighting. Several paths can be explored from this starting point, such as adding more data for the current languages in the setup, authentic or synthetic (e.g. via back-translation \cite{sennrich-etal-2016-improving}), or adding other languages that might share some common characteristics, like Hungarian for instance.

Some additional work could also be explored on the tokenization part. For simplicity, our first approach in this paper relies on a very simple shared BPE approach. But, some more sophisticated approaches, maybe language-specific or morphologically adapted \cite{micher-2018-using}, may be worth exploring.

Finally, some more novel approaches could be tried, like massive pre-training methods such as BART \cite{lewis2019bart}. A similar experimental process could be followed, starting from only the core languages of the task (English and Inuktitut), then extending to other languages and observe the impact.

\section{Conclusion}

Working on a new, unknown, language is always challenging. Even more so when this language is quite distant from any language you're used to. Also, automated metrics are far from being perfect for such tasks, especially in the context of such a particular language as Inuktitut.

Particularly for this task, human evaluation is key. But, as data, it's quite a scarce resource for Inuktitut. More knowledge of the language would be of tremendous help to better grasp the limits or interesting leads of the various models. One workaround can be to work on the opposite direction (Inuktitut to English), but there is no guarantee the model would have similar behaviour for similar tricks. And, some knowledge about Inuktitut would still be needed to analyze model behavior based on source inputs.

\bibliographystyle{acl_natbib}
\bibliography{emnlp2020}

\end{document}